\RequirePackage{fix-cm}
\documentclass[smallextended]{svjour3}       
\smartqed  
\usepackage{graphicx}
\usepackage{kotex}
\usepackage{color}
\usepackage{amsmath}
\usepackage{subcaption}
\usepackage{algorithm, algcompatible}
\usepackage{dingbat}

\newcommand{\etal}{\textit{\textit{et al.}}}
\newcommand{\ie}{\textit{i}.\textit{e}., }
\newcommand{\eg}{\textit{e}.\textit{g}., }

\begin{document}

\title{Style Transfer with Target Feature Palette and Attention Coloring
}


\author{Suhyeon Ha         \and
        Guisik Kim         \and
        Junseok Kwon 
}


\institute{Suhyeon Ha \at
              \email{tngus3752@gmail.com}           
           \and
           Guisik Kim \at
              \email{specialre@naver.com}
            \and
           Junseok Kwon \at
              \email{jskwon@cau.ac.kr}
            \at School of Computer Science and Engineering, Chung-Ang University, Seoul, Korea \\
}

\date{Received: date / Accepted: date}

\maketitle
\begin{abstract}
Style transfer has attracted a lot of attentions, as it can change a given image into one with splendid artistic styles while preserving the image structure. 
However, conventional approaches easily lose image details and tend to produce unpleasant artifacts during style transfer. 
In this paper, to solve these problems, a novel artistic stylization method with target feature palettes is proposed, which can transfer key features accurately.
Specifically, our method contains two modules, namely feature palette composition (FPC) and attention coloring (AC) modules. 
The FPC module captures representative features based on K-means clustering and produces a feature target palette. 
The following AC module calculates attention maps between content and style images, and transfers colors and patterns based on the attention map and the target palette. 
These modules enable the proposed stylization to focus on key features and generate plausibly transferred images.
Thus, the contributions of the proposed method are to propose a novel deep learning-based style transfer method and present target feature palette and attention coloring modules, and provide in-depth analysis and insight on the proposed method via exhaustive ablation study. 
Qualitative and quantitative results show that our stylized images exhibit state-of-the-art performance, with strength in preserving core structures and details of the content image.
\keywords{Non-Photorealistic Rendering (NPR)  \and Style transfer}
\end{abstract}

\begin{figure}[t]
\centering
\includegraphics[width=1.0\linewidth]{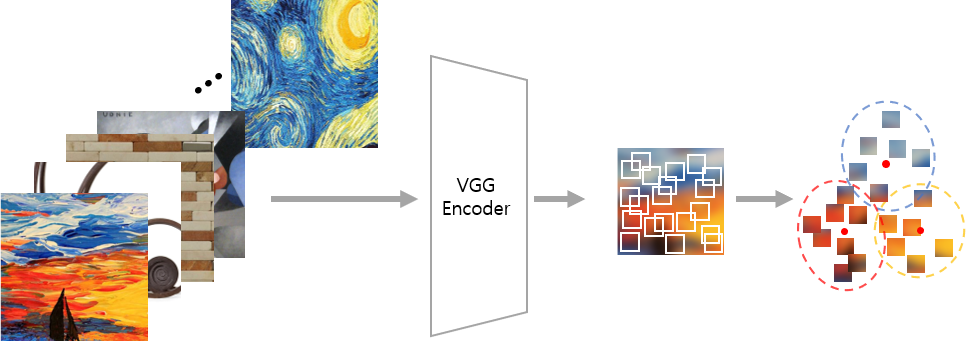}
\caption{\textbf{Illustration of the proposed style feature representation.}}
\label{fig:idea}
\end{figure}

\section{Introduction}
\label{intro}
Style transfer aims to transfer artistic patterns from given style images into content images, while preserving their structures. 
In particular, artistic stylization, which stems from non-photorealistic rendering in graphics, has been actively studied in the computer vision literature.

Recently, deep learning-based style transfer methods have achieved a great success owing to the seminal work of Gatys \etal{} \cite{gatys2016image}, which manipulates convolutional layers for rich feature representation.
To speed up the optimization procedure proposed by Gatys \etal, feed-forward network-based methods (\ie perceptual loss \cite{perceptual}, texture networks \cite{texturenet}, and improved texture networks \cite{improvedtexturenet}) have emerged.
One approach describes image styles using simple statistics (\eg AdaIN \cite{huang2017arbitrary} and WCT \cite{li2017universal}). 
Another approach represents the styles as relatively complex forms (\eg Gatys \etal{} \cite{gatys2016image} and CNNMRF \cite{li2016combining}).
However, the former has difficulty in preserving details of content images, whereas the latter produces spotty artifacts or duplicated textures of the style images. 

To solve the aforementioned problems, in this paper, a novel style transfer method with target feature palette is proposed, which uses a few distinctive features to match content and style images accurately.
These key style features are used for style transfer instead of using whole style image. 
As shown in Figure \ref{fig:idea}, style images are fed into the VGG encoder to obtain encoded style features. Then, these encoded features are cropped into small patches, which are grouped into three clusters.
These clusters are determined as distinct style features. We call them target feature palette.
The proposed method consists of two modules, which are feature palette composition (FPC) and attention coloring (AC) modules. 
The FPC module produces a target feature palette with representative features, which can be obtained using K-means clustering. 
The AC module facilitates stylization using the feature palette by recomposing style features according to the relationship between content and style images. 
Using FPC and AC, our method transfers major artistic attributes and preserves details of the content images, which produce high-quality stylized images.

Our contributions are summarized as follows:
 \begin{itemize}
     \item  We propose a novel deep learning-based style transfer method, which produces considerably diverse stylized images.
     \item  We present target feature palette and attention coloring modules, which especially have strength in maintaining detailed structures of content images.
     \item  To the best of our knowledge, we first adopt depth map errors as quantitative measures on large test data to compare the preservation of content's structures in style transfer.
     \item  We provide in-depth analysis on the proposed method via exhaustive ablation study. 
     To show the effectiveness of our method, we test the method according to the number of clusters, different sizes of patches, and multiple style images.
 \end{itemize}
 
 \begin{table*}[t]
 	\centering
 	\setlength{\tabcolsep}{16pt}
 	\caption{\textbf{Abbreviation.}}
 	\label{table:abbreviation}
 	\vspace{-1mm}
 	\begin{tabular}{|l|c|}
 		\hline
 		Abbreviation &  Description  \\ 
 		\hline
 		FPC  & Feature Palette Composition \\ 
 		AC   & Attention Coloring\\ 
 		NPR   & Non-Photorealistic Rendering\\ 
 		AdaIN   & Adaptive Instance Normalization\\ 
 		WCT   & Whitening and Coloring Transform\\ 
 		CNNMRF & Convolutional Neural Network Markov Random Field\\ 
 		GAN & Generative Adversarial Network\\ 
  		MST & Multimodal Style Transfer\\ 
    	AAMS & Attention-Aware Multi-Stroke\\ 
     	SANet & Style-Attentional Network\\ 
      	EFANet & Exchangeable Feature Alignment Network\\ 
      	AdaAttN & Attention Mechanism in Arbitrary Neural Style Transfer\\ 
      	DSTN & Domain-aware Style Transfer Networks\\
 		\hline
 	\end{tabular}
 \end{table*}
 \begin{table*}[t]
	\centering
	\setlength{\tabcolsep}{20pt}
	\caption{\textbf{Symbols.}}
	\label{table:symbol}
 	\vspace{-1mm}
	\begin{tabular}{|l|c|}
		\hline
		Abbreviation &  Description  \\ 
		\hline
		$I_c$  & Content image \\ 
		$I_s$  & Style image\\ 
		$F_c$   & Content features\\ 
		$F_s$   & Style features\\ 
		$m_{AdaIN}$   & Adaptive instance normalization module\\ 
		$m_{FPC}$ & Feature palette composition module\\ 
		$\mu$ & Mean\\ 
		$\sigma$ & Standard deviation\\ 
		$F_{cs}$ & Final stylized feature\\ 
		$I_{cs}$ & Final stylized image\\ 
		$\mathcal{L}_c$  & Content loss \\ 
		$\mathcal{L}_s$  & Style loss \\ 
		$E$ & Encoder\\ 
		$\phi_i$ & Feature map extracted from the $i$-th layer of VGG \\
		$L$ & A total number of layers\\ 
		$D_{c}$ & Depth map of content image\\ 
		$D_{cs}$ & Depth map of final stylized image\\ 
		$n$ & A total number of test pairs\\ 
		\hline
	\end{tabular}
\end{table*}
The remainder of the paper is organized as follows. 
Section \ref{sec:work} relates our method with existing approaches. 
In Section \ref{sec:proposed}, we explain the proposed FPC and AC modules for image stylization. 
We qualitatively and quantitatively compare the proposed method with state-of-the-art methods and various experiments of style transfer in Section \ref{sec:exp} and draw a conclusion in Section \ref{sec:con}.
Tables \ref{table:abbreviation} and \ref{table:symbol} summarize abbreviations and symbols, respectively.

\section{Related Work} \label{sec:work}

\begin{table*}[t]
 	\centering
 	\setlength{\tabcolsep}{5pt}
 	\caption{\textbf{{Features on style transfer methods.}}}
 	\label{table:related}
 	\vspace{-3mm}
 	\begin{tabular}{|c|c|c|c|c|}
 		\hline
 		Year & Title & Content & Color & Pattern \\
 		 & & Preservation & Reproduction & Reproduction\\
 		\hline
 		2016 & Gatys \etal \cite{gatys2016image} & & \checkmark &\\
 		\hline
 		2017 & WCT \cite{li2017universal} & \checkmark & \checkmark &\\
 		\hline
 		2017 & AdaIN \cite{huang2017arbitrary} & \checkmark & \checkmark & \\
 		\hline
 		2018 & Avatar-Net \cite{sheng2018avatar} & & \checkmark & \checkmark \\
 		\hline
 		2019 & SANet \cite{park2019arbitrary} & & \checkmark & \checkmark  \\
 		\hline
 		2021 & DSTN \cite{DSTN} & & \checkmark & \checkmark  \\
 		\hline
 	\end{tabular}
 \end{table*}
 
Table \ref{table:related} summarizes the pros and cons of conventional methods.

\subsection{Texture Synthesis}
Texture synthesis has emerged to transform random noise into textures in a given image.
Conventional texture synthesis methods can be categorized into two groups, parametric (\eg pyramid-based texture analysis/synthesis \cite{pyramid-texture} and parametric texture models based on joint statistics of complex wavelet coefficients \cite{param-texture}) and non-parametric (\eg synthesis using neighboring information \cite{efros1999texture,wei2000fast,efros2001image,liang2001real}). 
The former employs statistical models to describe textures through the visual matching, whereas the latter synthesizes images by sampling similar pixels \cite{efros1999texture,wei2000fast} or patches \cite{efros2001image,liang2001real}.

\subsection{Optimization-based Style Transfer}
Regarding style transfer, Gatys \etal~presented the seminal work \cite{gatys2016image}, in which a white noise is changed into an impressive artwork.
In this work, Gatys \etal~used gram matrices obtained by channel-wise correlation of extracted style features from each layer of pre-trained deep neural networks.
CNNMRF \cite{li2016combining} adopted a generative MRF model on top of VGG networks and increased visual plausibility.
However, they require complex optimization to generate synthesized images, which inevitably induces high computational costs.

\subsection{Feed-forward Networks-based Style Transfer }
To alleviate this problem, several feed-forward networks have been proposed in \cite{ulyanov2016texture,MGAN,perceptual}.
Ulaynov \etal{} \cite{ulyanov2016texture} introduced multi-scale feed-forward generation networks.
Li and Wand \cite{MGAN} proposed Markovian generative adversarial networks.
Johnson \etal{} \cite{perceptual} replaced a per-pixel loss with a perceptual loss to solve the optimization problems.
However, these methods allow the networks to learn only a specific style. 
Thus, a whole network has to be retrained to learn a new style.

\subsection{Arbitrary Style Transfer }
Meanwhile, multiple styles has been considered for arbitrary style transfer \cite{dumoulin2016learned,chen2016fast,huang2017arbitrary,li2017universal,sheng2018avatar,park2019arbitrary}.
For example, Dumoulin \etal{} \cite{dumoulin2016learned} introduced conditional instance normalization and learned a different set of affine parameters for each style with the same convolutional parameters.
Style Swap \cite{chen2016fast} performed a stylization process by exchanging patches in the feature space.
Huang and Belongie \cite{huang2017arbitrary} proposed adaptive instance normalization(AdaIN) to compute a set of affine parameters from a given style image.
WCT \cite{li2017universal} transformed content features to the covariance matrix of style features through whitening and coloring steps. 
Avatar-Net \cite{sheng2018avatar} proposed a multi-scale style transfer method based on the patch-based feature manipulation.
SANet \cite{park2019arbitrary} synthesized style images in real-time using style-attentional networks.

\subsection{Recent Style Transfer}
Recently, generative adversarial networks (GANs) \cite{goodfellow2014generative} have been used not only to solve a variety of vision problems (\eg scene classification \cite{sceneclassification} and co-saliency detection \cite{cosaliency}) but also to generate synthesized images.
Zhu \etal{} \cite{zhu2017unpaired} transformed style transfer into unsupervised image-to-image translation problems and presented a novel GAN architecture with the cycle consistency loss and the bijection operation between two image domains.
Tomei \etal{} \cite{tomei2019art2real} translated paintings into photo-realistic images using memory banks in a semantically-aware way.
Although they produced impressive results, they more focus on the mapping between two image collections rather than the mapping between two specific images, which is explicitly considered as the content and style images in our method.
The latest style transfer works continued to make progress in various ways.
Deng \etal{} \cite{multi-adapt} and Wu \etal{} \cite{efanet} used disentanglement approaches to split complex features into manipulable forms such as content, style, and common features and let features interact with each other.
Moreover, Wang \etal{} \cite{dfp} presented feature perturbation which can increase a diversity adding to previous WCT-based style transfer methods.
Zhang \etal{} \cite{msgnet} solved a style transfer problem by learning differentiable second-order statistics.
Wang \etal{} \cite{ultra-resolution} applied knowledge distillation and dealt with ultra-resolution sized images.

\subsection{Content Preservation in Style Transfer }

Many conventional style transfer methods tend to fail in retaining the content structure over the course of style transfer due to texture transfer. They easily generate spotty artifacts or duplicated texture preventing from recognizing objects in stylized images. For all that, style transfer has had fewer interests in how to generate images maintaining the content. Cheng \etal{} \cite{cheng2019structure} first utilized depth map and edge detection as guidance to retain the structure of content images and improve the quality of style transfer. To solve the sill remaining problems, in this paper, a novel style transfer method with a target feature palette is proposed, which uses a few distinctive features to match content and style images accurately, thus maintaining content after the transition.

\begin{figure*}[t]
    \begin{minipage}[b]{1.0\linewidth}
        \centering
        \includegraphics[width=1.0\linewidth]{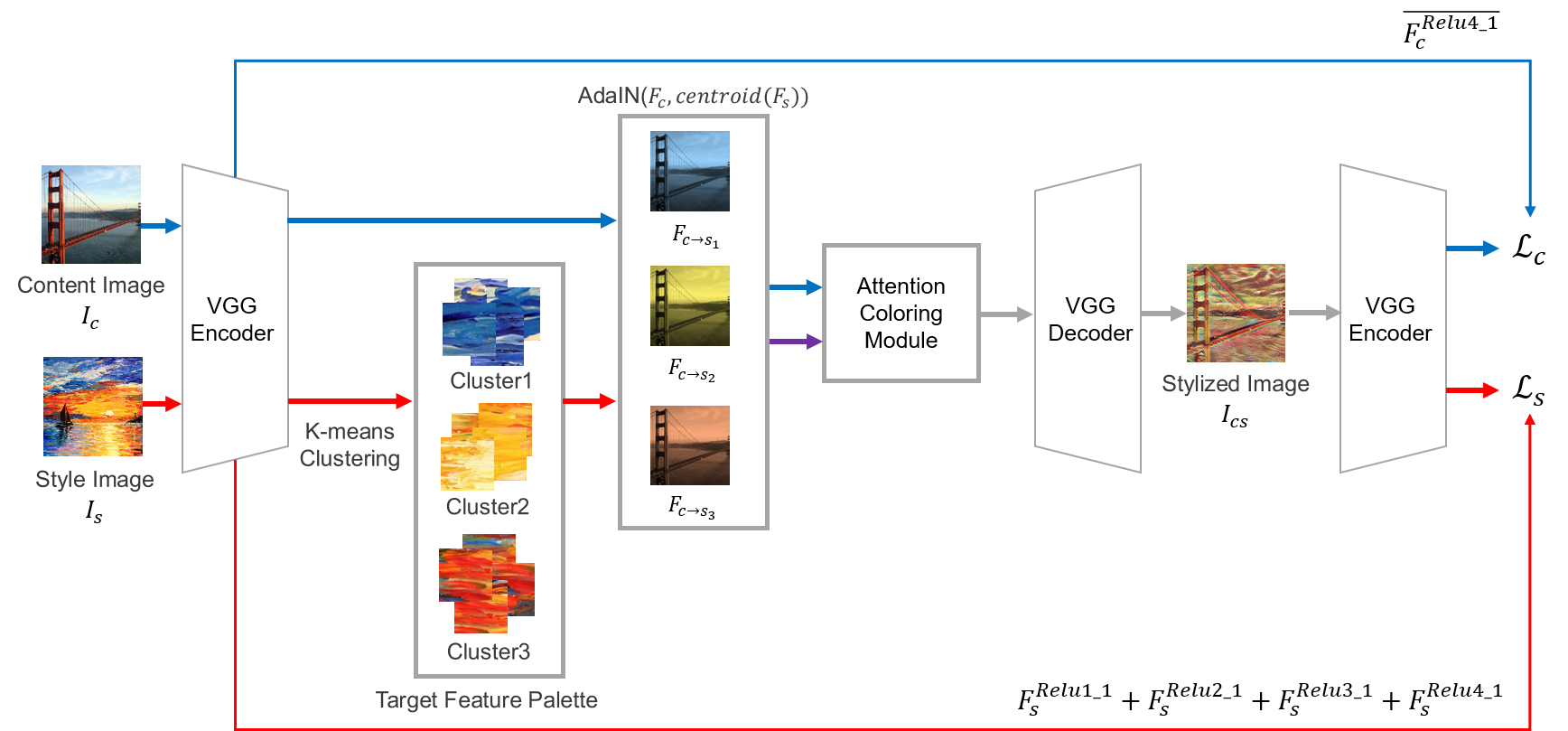}
    \end{minipage}
  \caption{\textbf{Overall procedure of the proposed framework.}}
 \label{fig:network}
\end{figure*}
\begin{algorithm}[t]
    \caption{TFP-AC Style Transfer}
    \label{eq:algorithm}
    \begin{algorithmic}[1]
        \REQUIRE A pair of content image $I_c$ and style image $I_s$
        \ENSURE A stylized image $I_{cs}$
		\FOR {$I_c$ and $I_s$}
		\STATE Extract content features $F_c$ and style features $F_s$ using pre-trained VGG-19
		\STATE K-means clustering with the number of $k$ using $F_s$
		\FOR {Each cluster}
		\STATE Find the centroid patch denoted as $m_{FPC}(F_s)$
		\STATE Stylization $F_c$ with $m_{FPC}(F_s)$ using AdaIN (Eq.: 1, 2)
		\ENDFOR
		\FOR {Each cluster}
		\STATE Mean-variance channel-wise normalization of $F_c$ and $F_{c \rightarrow s_k}$
		\STATE Estimation of attention map
		\STATE Multiplication of attention map and $F_{c \rightarrow s_k}$
		\ENDFOR
		\STATE Sum over all $\hat{F}_{c \rightarrow s_k}$ in element-wise manner
		\STATE $3\times3$ convolution over the summation
		\STATE Feed $F_{cs}$ to decoder
		\ENDFOR
	\end{algorithmic} 
\end{algorithm} 

\section{Proposed Method}  \label{sec:proposed}
The proposed deep neural network adopts a encoder-decoder architecture consisting of two modules: FPC and AC modules, as shown in Figure \ref{fig:network}. 
Content and style images are fed into the FPC module to obtain a target feature palette. Then, images are stylized using AdaIN and AC module, where $\overline{F_c^{\texttt{Relu4\_1}}}$ denotes the normalized content feature extracted from \texttt{Relu4\_1} of the VGG encoder. 
The overall procedure (pseudocode) is described in Algorithm \ref{eq:algorithm}.
 
\subsection{Feature Palette Composition Module}
Given content image $I_c$ and style image $I_s$, content features $F_c$ and style features $F_s$ are extracted using the VGG-19 pre-trained encoder in \cite{simonyan2014very}. 
Then, the FPC module splits $F_s$ into twenty patches with the size of $8\times8$ and applies K-means clustering to extract $k$ (the number of clusters) distinctive groups having similar features. 
We use a centroid of each cluster for composing a target palette instead of using a centroid-nearest patch for experiments in Section \ref{sec:exp}.
Subsequently, we conduct the first stylization by converting the content features into stylized features of the target palette using its mean and standard deviation, which can be implemented by using adaptive instance normalization $m_{AdaIN}(\cdot,\cdot)$ \cite{huang2017arbitrary}, as follow:
\begin{equation}\label{eq:fpc_equation}
     m_{AdaIN}(F_c, m_{FPC}(F_s)),
\end{equation}
where $m_{FPC}(\cdot)$ indicates the FPC module. 
In \eqref{eq:fpc_equation}, $n(\cdot,\cdot)$ is defined as 
\begin{equation}\label{eq:adain_equation}
     m_{AdaIN}(x, y) = \sigma(y)\left(\frac{x-\mu(x)}{\sigma(x)}\right)+\mu(y),
\end{equation}
where $\mu(x)$ and $\sigma(x)$ denote the mean and standard deviation of $x$, respectively.
Using three distinct styles of the target feature palette (\ie $s_1$, $s_2$, and $s_3$), $F_c$ is converted into $F_{c \rightarrow s_1}$, $F_{c \rightarrow s_2}$, and $F_{c \rightarrow s_3}$, respectively. 
Then, $F_{c \rightarrow s_1}$, $F_{c \rightarrow s_2}$, $F_{c \rightarrow s_3}$ and $F_c$ are fed into the following AC module $m_{AC}(\cdot,\cdot)$:
\begin{equation}\label{eq:whole_equation}
     m_{AC}\Big(F_c, m_{AdaIN}\big(F_c,m_{FPC}(F_s)\big)\Big),
\end{equation}
where $m_{AC}(\cdot,\cdot)$ is explained in the next section. 
\begin{figure}[t]
\centering
\includegraphics[width=1.0\linewidth]{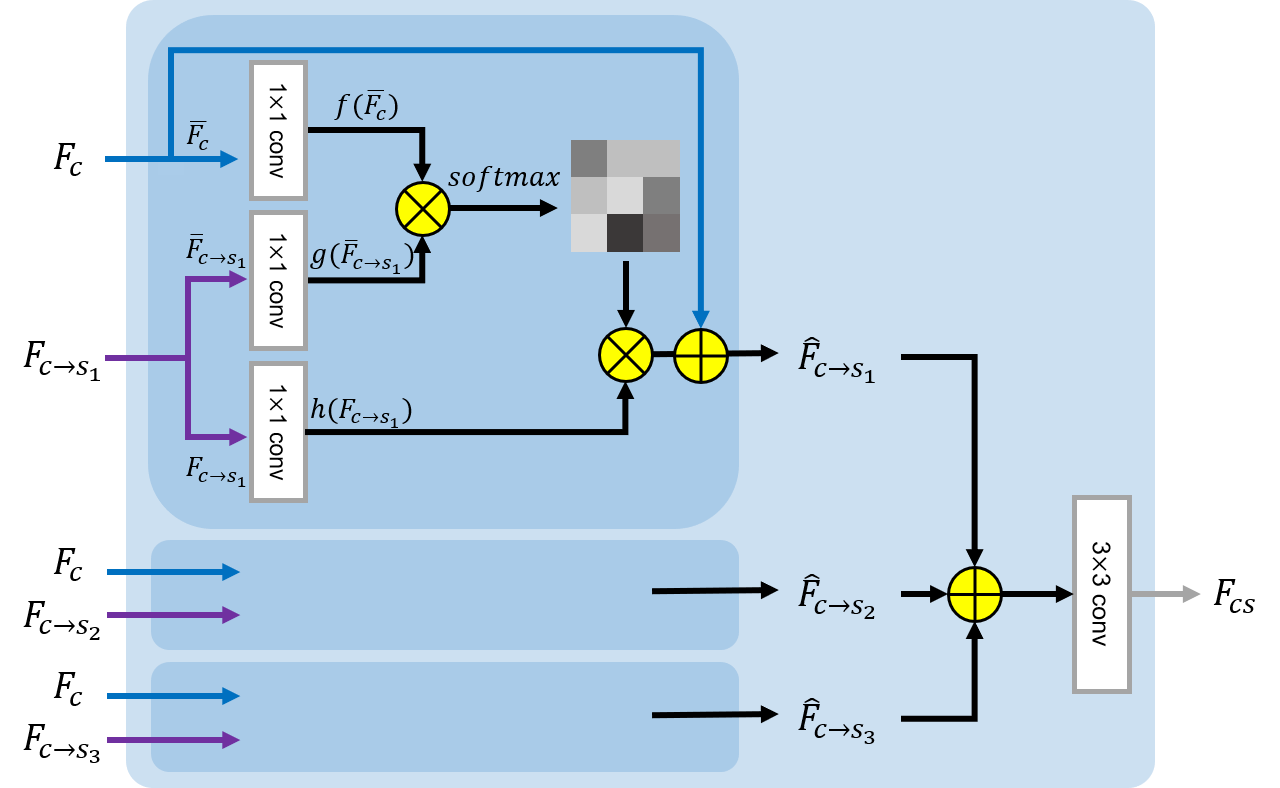}
\caption{\textbf{Attention Coloring Module.} The attention map between content and style features is used to reassemble features of stylized image.}
\label{fig:AC}
\end{figure}
\subsection{Attention Coloring Module}
The AC module estimates the attention map and reassemble the features of stylized images $F_{c \rightarrow s_k}$ into attentive stylized images $\hat{F}_{c \rightarrow s_k}$, as shown in Figure \ref{fig:AC}. 
In particular, $F_c$ and $F_{c \rightarrow s_k}$ are mean-variance channel-wise normalized into two feature spaces $f$ and $g$, which produces $f(\overline{F}_c)$ and $g(\overline{F}_{c \rightarrow s_k})$, respectively.
Then, the attention map is estimated by applying $softmax$ to the multiplication of $f(\overline{F}_c)$ and $g(\overline{F}_{c \rightarrow s_k})$.
To reassemble original stylized images, $F_{c \rightarrow s_k}$ is mapped into the feature space $h$ and is multiplied by the attention map, which results in $\hat{F}_{c \rightarrow s_k}$. 
All $\hat{F}_{c \rightarrow s_k}$ are summed in an element-wise manner and processed with $3\times3$ convolution, which produces $F_{cs}$. 
The final stylized image $I_{cs}$ can be obtained by feeding $F_{cs}$ into the decoder.

The proposed AC module stemming from SANet \cite{park2019arbitrary} produces more confidential stylized results with detailed structures by combining all $\hat{F}_{c \rightarrow s_k}$.
(1) the inner part strengthens the content, because calculating the spatial attention map between content feature $F_{c}$ and stylized feature $F_{c \rightarrow s_k}$ is highly related to the structure of given content.
Note that $F_{c \rightarrow s_k}$ is originated from content images but have transferred styles via \textit{AdaIN}.
(2) the outer part also intensifies the content in the course of combining all $F_{c \rightarrow s_k}$ into $F_{cs}$.
This is because each $F_{c \rightarrow s_k}$ that highlights its own style $s_k$ is also accumulated.
This content-focused mechanism of the inner and outer procedure allows the stylized images to retain detailed structures, while traditional methods fail to keep structures in a situation of content-style trade-off.

\subsection{The Loss Function}
Similar to existing approaches in \cite{chen2016fast,huang2017arbitrary,li2017universal,sheng2018avatar}, we use the layers from \texttt{Relu1\_1} to \texttt{Relu4\_1} of the pre-trained VGG-19 network for the encoder. 
The randomly initialized decoder mostly mirrors the encoder. 
The loss function is a weighted combination of the content loss $\mathcal{L}_c$ and the style loss $\mathcal{L}_s$ with weights $\lambda_c$ and $\lambda_s$, respectively:
\begin{equation}\label{eq:full_loss}
    \mathcal{L} = \lambda_c\mathcal{L}_c+\lambda_s\mathcal{L}_s,
\end{equation}
where $\lambda_c$ and $\lambda_s$ are set to $30$ and $1$, respectively. 
The content loss is measured by the Euclidean distance between mean-variance channel-wise normalized feature of the output image $\overline{E(I_{cs})^{\texttt{Relu4\_1}}}$ and the mean-variance channel-wise normalized target feature $\overline{F_c^{\texttt{Relu4\_1}}}$:
\begin{equation}\label{eq:content_loss}
	\mathcal{L}_c = \left\Vert\overline{E(I_{cs})^{\texttt{Relu4\_1}}}-\overline{F_c^{\texttt{Relu4\_1}}}\right\Vert_2.
\end{equation}
where $E(\cdot)$ denotes the encoder and \texttt{Relu4\_1} is the VGG feature map obtained from the corresponding layer.
The style loss is defined as follows:
\begin{equation}\label{eq:style_loss}
    \begin{split}
        \mathcal{L}_s = \sum_{i=1}^{L} \left\Vert\mu(\phi_i(I_{cs}))-\mu(\phi_i(I_s))\right\Vert_2 \\
                        + \left\Vert\sigma(\phi_i(I_{cs}))-\sigma(\phi_i(I_s))\right\Vert_2,
    \end{split}
\end{equation}
where $\phi_i$ is the feature map extracted from the $i$-th layer of VGG and $L$ denotes a total number of layers.
Conventional methods including \cite{gatys2016image} and \cite{perceptual} use values of covariance of style features, called the Gram matrix.
In contrast, our method makes use of style features itself to minimize the difference between style images and stylized images.

 \vspace{3mm}
\section{Experimental Results} \label{sec:exp}

\begin{figure*}[t]
    \begin{minipage}[b]{1.0\linewidth}
        \centering
        \includegraphics[width=0.9\linewidth]{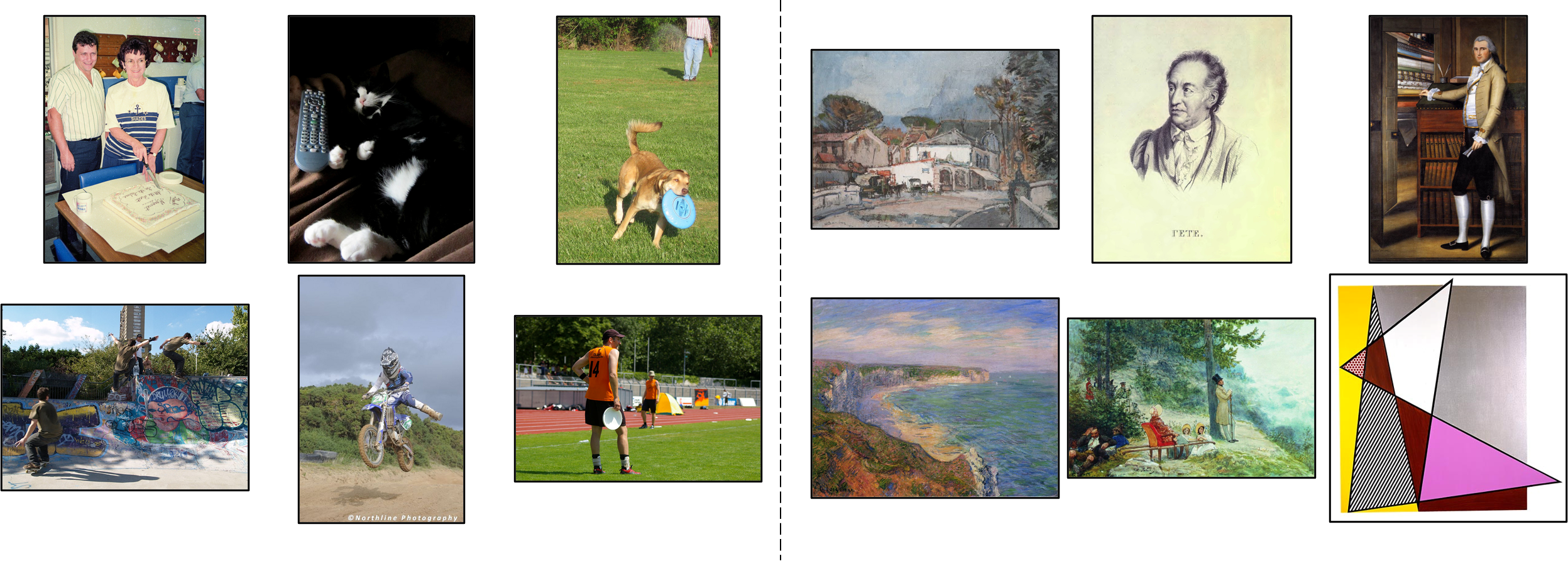}
    \end{minipage}
    \begin{minipage}[b]{0.5\linewidth}
		\centering
		(a) MS-COCO (Content)
	\end{minipage}
	\begin{minipage}[b]{0.5\linewidth}
		\centering
		(b) Painter by Numbers (Style)
	\end{minipage}
  \caption{\textbf{Sample images from content and style datasets.}}
 \label{fig:dataset}
\end{figure*}

\begin{table*}[t]
 	\centering
 	\setlength{\tabcolsep}{20pt}
 	\caption{\textbf{Training configuration.}}
 	\label{table:setting}
 	\vspace{-3mm}
 	\begin{tabular}{|c|c|}
 		\hline
 		\multicolumn{2}{|c|}{Experimental settings} \\
 		\hline
 		Input size & $512\times512$ \\
 		\hline
 		Batch size & $4$ \\
 		\hline
 		Total iterations & $50,000$ \\
 		\hline
 		Optimizer & Adam optimizer \\
 		\hline
 		Learning rate & $0.0001$ \\
 		\hline
 		Patch size & $8$ \\
 		\hline
 		Cluster size & $3$ \\
 		\hline
 		The number of patches & $20$ \\
 		\hline
 	\end{tabular}
 \end{table*}

\subsection{Implementation Details}

We trained the proposed network using MS-COCO \cite{lin2014microsoft} and Painter by Numbers \cite{nichol2016painter} (PBN) for content and style datasets, respectively. MS-COCO includes 1.5 million object categories. Owing to its magnitude, it is widely used for object recognition and pre-training for many vision tasks. PBN is art images collected from WikiArt ranging from still life, calligraphy to abstract in the genre, from the second century to the 21st century in date. Randomly selected sample images from each dataset are shown in Figure \ref{fig:dataset}. Each dataset contains approximately $80,000$ images.

Table \ref{table:setting} lists configuration during training. Each element was selected by considering the stability of learning and content preservation. A variety of experiments on different settings (\eg representative patch, patch size, and cluster size) are executed in Section \ref{sec:ablation}. We used the Adam optimizer \cite{kingma2014adam} with a learning rate of $0.0001$ and a batch size of $4$ for content-style image pairs. We trained our network for $50$k iterations.
During training, we resized images to $512$ while preserving the aspect ratio, then randomly cropped regions with the size of $256\times256$. 
We used twenty randomly cropped patches with the size of $8\times8$ for clustering in training after encoding. In testing, we resized images to $512$ and utilized a hundred randomly cropped patches with the size of $8\times8$.
The number of clusters was fixed to $3$ for training and testing.
We performed the experiments on a workstation with an NVIDIA GeForce RTX 2080Ti GPU.
The proposed network was implemented using Pytorch.
We set content and style weights, $\lambda_c$ and $\lambda_s$, to $30$ and $1$. In some parts of the ablation study, we slightly adjust the weights.

\begin{figure}[t]
    \centering
    \begin{subfigure}[b]{0.49\textwidth}
        \includegraphics[width=\textwidth]{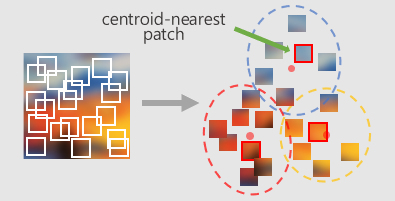}
        \caption{Patch-based FPC module}
    \end{subfigure}
    \begin{subfigure}[b]{0.49\textwidth}
        \includegraphics[width=\textwidth]{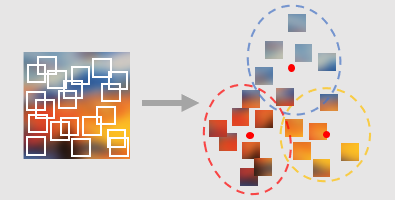}
        \caption{Centroid-based FPC module}
        \label{fig:tiger}
    \end{subfigure}
    \caption{\textbf{Two variants of the proposed FPC module}.}
    \label{fig:patch_centroid}
\end{figure}
\begin{figure}[t]
    \begin{minipage}[b]{1.0\linewidth}
        \centering
        \includegraphics[width=1.0\linewidth]{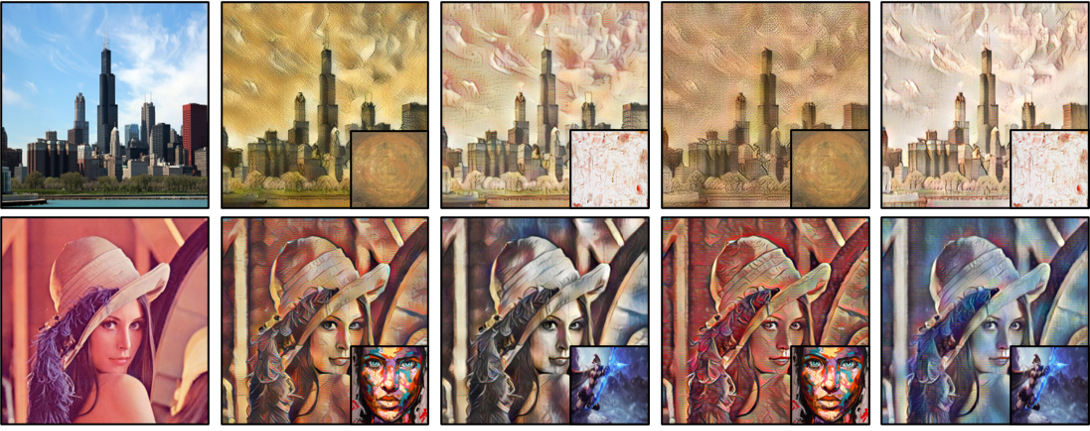}
    \end{minipage}
	\begin{minipage}[b]{0.19\linewidth}
		\centering
		(a) Content images
	\end{minipage}
	\begin{minipage}[b]{0.38\linewidth}
		\centering
		(b) Results using patch-based FPC module
	\end{minipage}
	\begin{minipage}[b]{0.38\linewidth}
		\centering
		(c) Results using centroid-based FPC module
	\end{minipage}
  \caption{\textbf{Style transfer results using patch and centroid-based FPC modules.} Style images are shown in right-bottom boxes at each column.}
 \label{fig:vs}
\end{figure}
\begin{figure}[t]
    \begin{minipage}[b]{1.0\linewidth}
        \centering
        \includegraphics[width=1.0\linewidth]{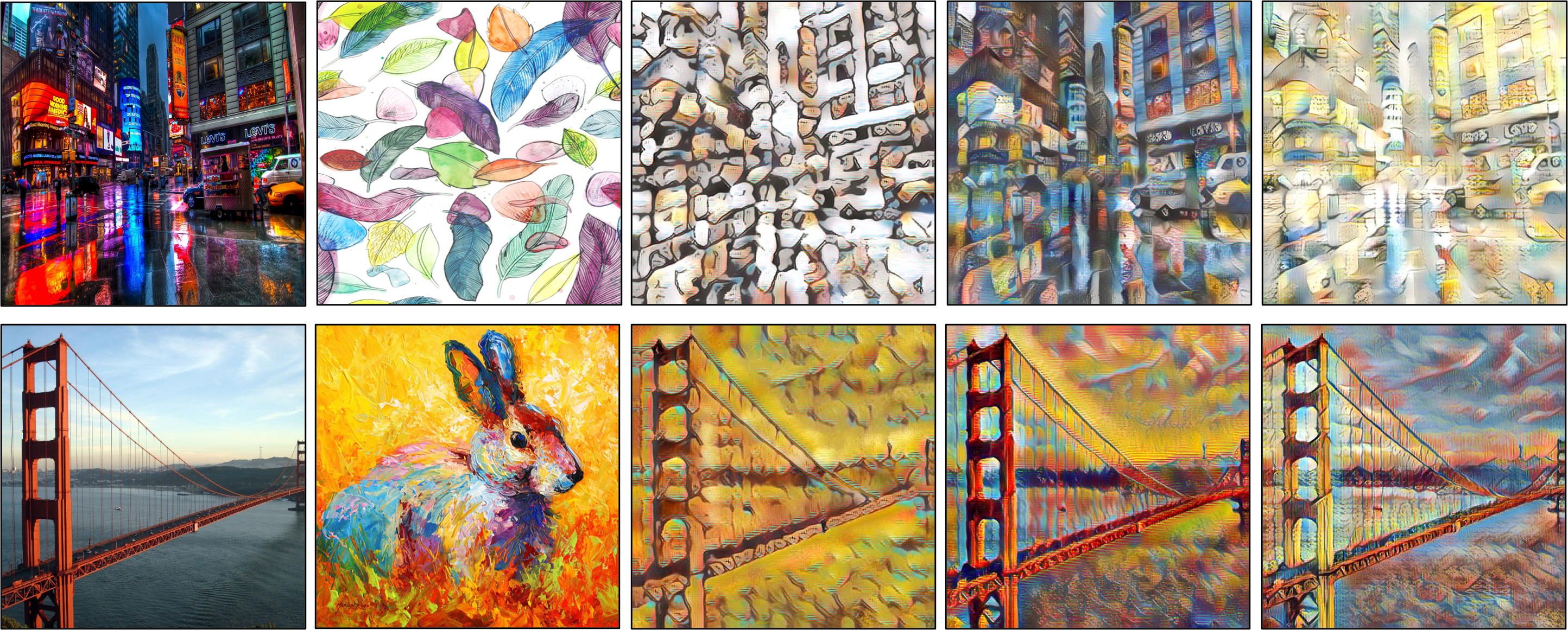}
    \end{minipage}
    \begin{minipage}[b]{0.19\linewidth}
		\centering
		(a) Content images
	\end{minipage}
	\begin{minipage}[b]{0.19\linewidth}
		\centering
		(b) Style \\images
	\end{minipage}
	\begin{minipage}[b]{0.19\linewidth}
		\centering
		(c) Stylized with $k=1$
	\end{minipage}
	\begin{minipage}[b]{0.19\linewidth}
		\centering
		(d) Stylized with $k=3$
	\end{minipage}
	\begin{minipage}[b]{0.19\linewidth}
		\centering
		(e) Stylized with $k=5$
	\end{minipage}
  \caption{\textbf{Style transfer results according to the number of clusters.} The result differs with the size of $k$ given the same inputs.}
 \label{fig:k}
\end{figure}

\subsection{Ablation Study} \label{sec:ablation}
To provide more in-depth analysis of and insight into the proposed method, we performed ablation experiments for the individual components, \ie FPC and AC. 
Our method can diversify the stylized image by the help of the FPC module and produce multi-style transfer results using the AC module.

\subsubsection{Effectiveness of the proposed FPC module} 
For this experiment, we made two variants of the FPC module: patch-based and centroid-based. 
In the patch-based FPC module, we selected a centroid-nearest patch in each cluster as the representative patch, as shown in Figure \ref{fig:patch_centroid}(a).
In the centroid-based FPC module, we used a centroid value of the style features for each cluster, as shown in Figure \ref{fig:patch_centroid}(b).   
Figure \ref{fig:vs} compares stylized image results using the centroid-based FPC module with those using the patch-based FPC module. 
As shown in Figure \ref{fig:vs}, the centroid-based FPC module produces qualitatively better results than the patch-based FPC module. 
For example, in the second-fourth columns of the second row, centroid-based results reflect a diversity of face colors in the style images. 
In the third-fifth columns of the second row, centroid-based results fully represent blue colors, which are distinct features of the style images. 
The centroid-based module showed good performance, because a centroid is an intermediate value between style features and it contains more fruitful feature information than using a single feature. 
These results implies that our clustering procedure is helpful for extracting representative features and accurately reconstructing styles.

Figure \ref{fig:k} shows style transfer results according to the number of clusters, $k$. To verify the effect of the number of clusters, each model was trained and tested by changing the number of clusters $k = 1$, $3$, or $5$ respectively. When $k=1$(c), although the model can extract the main color with the clustering, it cannot hold the basic structure without combining procedures. When $k=3$(d), stylized images display the various colors shown in the style images without changing the content image extremely. The results from the setting of $k=5$(e) also contain multiple colors but the model tends to make a patch group including backgrounds, which leads to output blurry and less vivid images. Therefore, we set $k$ to $3$ for both the training and testing empirically.

\begin{figure}[t]
    \begin{minipage}[b]{1.0\linewidth}
        \centering
        \includegraphics[width=1.0\linewidth]{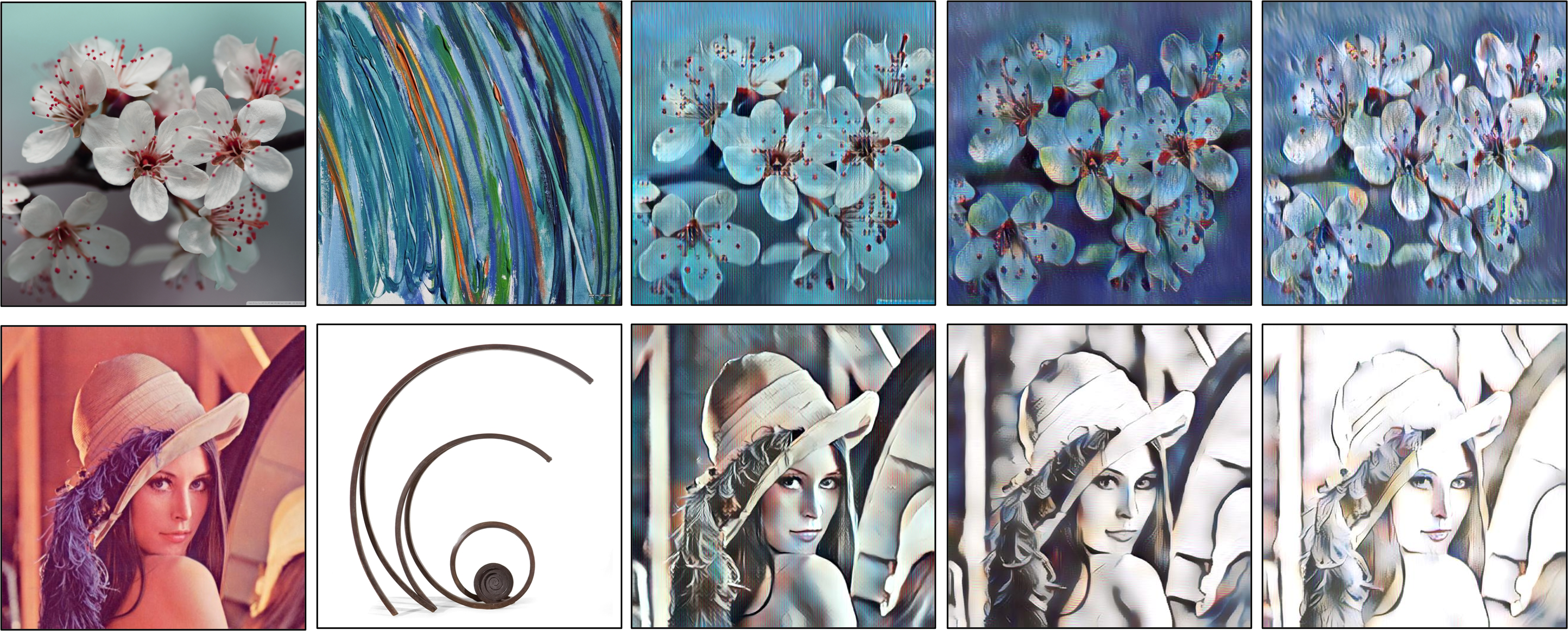}
    \end{minipage}
    \begin{minipage}[b]{0.19\linewidth}
		\centering
		(a) Content images
	\end{minipage}
	\begin{minipage}[b]{0.19\linewidth}
		\centering
		(b) Style \\images
	\end{minipage}
	\begin{minipage}[b]{0.19\linewidth}
		\centering
		(c) Stylized with $p=4$
	\end{minipage}
	\begin{minipage}[b]{0.19\linewidth}
		\centering
		(d) Stylized with $p=8$
	\end{minipage}
	\begin{minipage}[b]{0.19\linewidth}
		\centering
		(e) Stylized with $p=16$
	\end{minipage}
  \caption{\textbf{Style transfer results according to the size of patches.} }
 \label{fig:p}
\end{figure}

In Figure \ref{fig:p}, two pairs of the image were stylized with different patch sizes $p = 4$, $8$, and $16$. Note that the referred patch size is obtained after encoding, thus it is equal to a quarter of the size in three dimensions. As shown in the first row, the bigger is the size of patches, the more expressive is the texture of a given style image (\eg vertical pattern). However, the larger patch is likely to prevent capturing small but distinctive features (\eg solid brown line) as shown in the second row. For these reasons, we set $p$ to $8$ for our training and testing.

\begin{figure}[t]
    \begin{minipage}[b]{1.0\linewidth}
        \centering
        \includegraphics[width=1.0\linewidth]{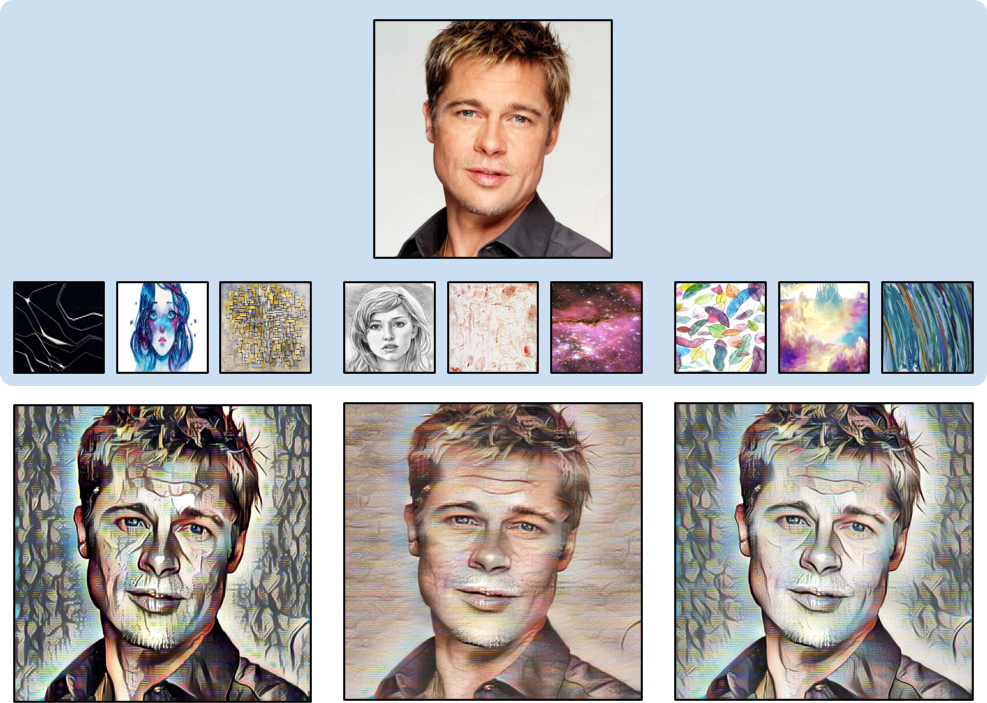}
    \end{minipage}
  \caption{\textbf{Multi-style transfer results.} }
 \label{fig:multi}
\end{figure}

\begin{figure}[t]
    \centering
    \begin{subfigure}[b]{1.0\textwidth}
        \includegraphics[width=\textwidth]{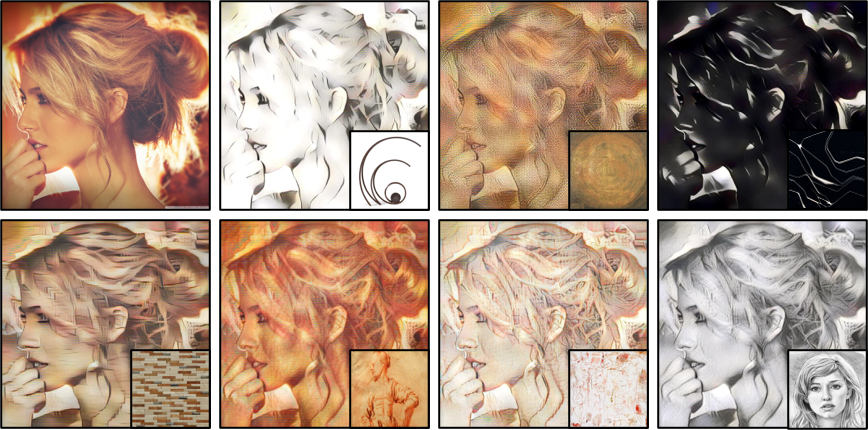}
    \end{subfigure}
    \begin{minipage}{1.0\linewidth}
		\centering
		{(a) Diverse stylized results 1}
	\end{minipage}
    \begin{subfigure}[b]{1.0\textwidth}
        \includegraphics[width=\textwidth]{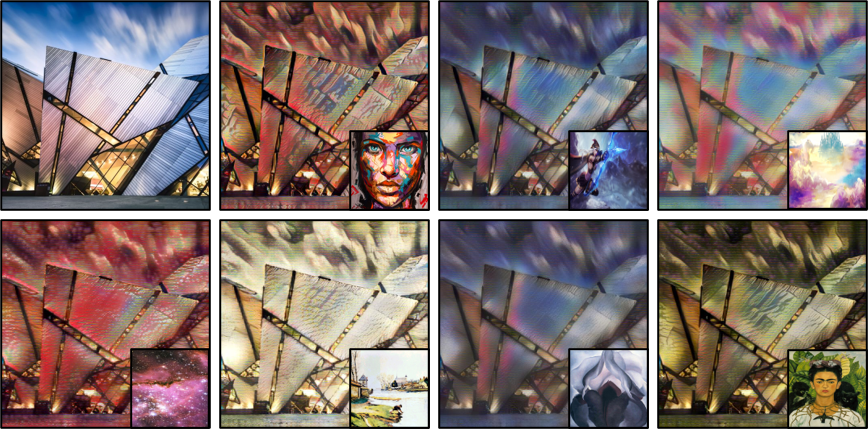}
    \end{subfigure}
    \begin{minipage}{1.0\linewidth}
		\centering
		{(b) Diverse stylized results 2}
	\end{minipage}
\caption{\textbf{Diverse stylized results.} }
\label{fig:arbitrary}
\end{figure}

\subsubsection{Effectiveness of the proposed AC module}
Figure \ref{fig:multi} contains multi-style transfer results, which are obtained using multiple style images simultaneously for single content image.
For the multi-style transfer experiment, we used one content image and three style images.
In particular, the input images are shown in the blue box, while three multi-stylized images are generated using four input images (\ie one content image and three style images) with different settings.
For simplicity, we randomly chose one centroid from a target feature palette from each style image. In Figure \ref{fig:multi}, experimental results can be different according to choices of target palette. Various compositions of palette result in stylization diversity.
With respect to style mixing, AC module refers to feature fusion of style images based on the feature palette. For example, in the first setting, three input style images have dark tone, blue color in cartoon style and repetitive patterns respectively. The first stylized image shows dark and blue color over the image and continuous patterns around a figure.
Similarly, in the second setting, light sketch and pink color of style images are well described in the stylized image.
Therefore, proposed AC module adaptively fused multiple style features based on the attention map and produced qualitatively accurate stylized images.

\begin{figure}[t]
\centering
\includegraphics[width=1\linewidth]{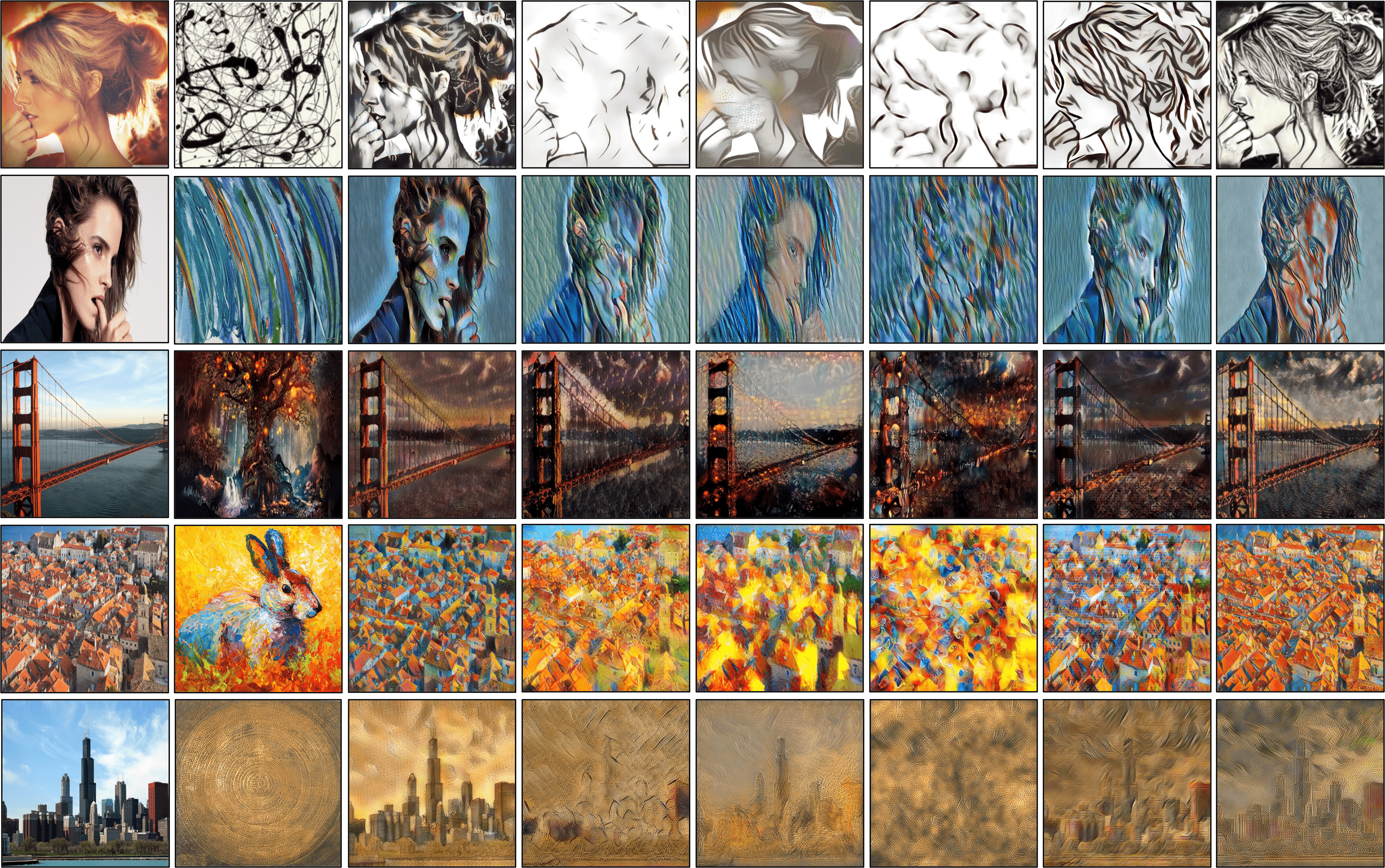}
    \begin{minipage}{0.12\linewidth}
      \centering
        {(a) \\Content}
    \end{minipage}  
    \begin{minipage}{0.12\linewidth}
      \centering
        {(b) \\Style}
    \end{minipage} 
    \begin{minipage}{0.12\linewidth}
      \centering
        {(c) \\Ours}
    \end{minipage} 
    \begin{minipage}{0.11\linewidth}
      \centering
        {(d) \\AdaIN\\\cite{huang2017arbitrary}}
    \end{minipage} 
    \begin{minipage}{0.11\linewidth}
      \centering
        {(e) \\Gatys\\\etal{} \cite{gatys2016image}}
    \end{minipage} 
    \begin{minipage}{0.12\linewidth}
      \centering
        {(f) \\WCT \cite{li2017universal}}
    \end{minipage} 
    \begin{minipage}{0.12\linewidth}
      \centering
        {(g) \\MST \cite{zhang2019multimodal}}
    \end{minipage} 
    \begin{minipage}{0.12\linewidth}
      \centering
        {(h) \\AdaAttN \\\cite{adaattn}}
    \end{minipage} 
  \caption{\textbf{Qualitative comparison with state-of-the-art methods.}}
\label{fig:comparison1}
\end{figure}
\begin{figure}[t]
\centering
\includegraphics[width=1\linewidth]{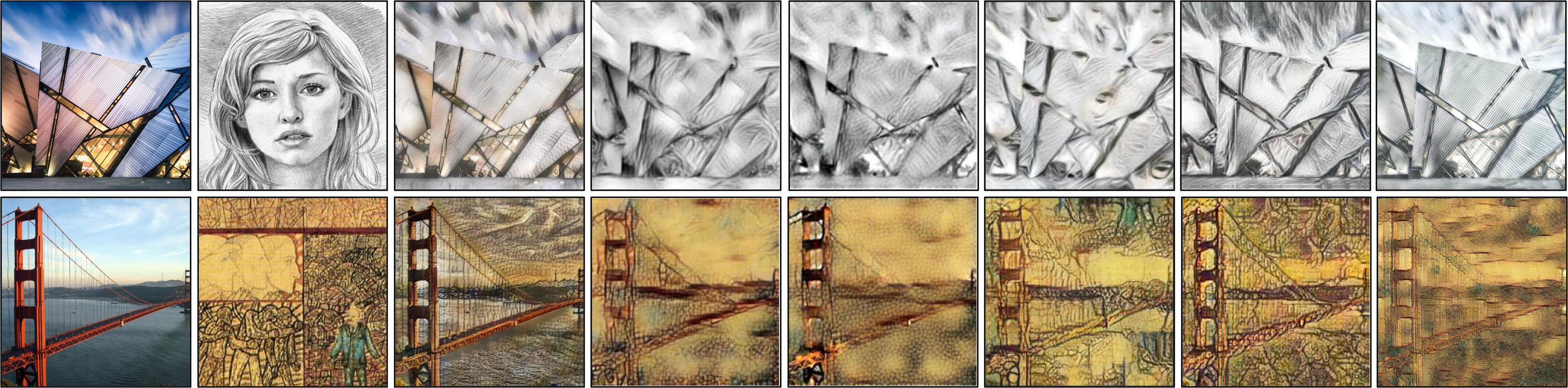}
    \begin{minipage}{0.12\linewidth}
      \centering
        {\footnotesize{(a) \\Content}}
    \end{minipage}  
    \begin{minipage}{0.12\linewidth}
      \centering
        {(b) \\Style}
    \end{minipage} 
    \begin{minipage}{0.12\linewidth}
      \centering
        {(c) \\Ours}
    \end{minipage} 
    \begin{minipage}{0.11\linewidth}
      \centering
        {(d) \\Avatar\\-Net \cite{sheng2018avatar}}
    \end{minipage} 
    \begin{minipage}{0.11\linewidth}
      \centering
        {(e) \\AAMS \cite{aams}}
    \end{minipage} 
    \begin{minipage}{0.12\linewidth}
      \centering
        {(f) \\SANet \\\cite{park2019arbitrary}}
    \end{minipage} 
    \begin{minipage}{0.12\linewidth}
      \centering
        {(g) \\EFANet \cite{efanet}}
    \end{minipage} 
    \begin{minipage}{0.12\linewidth}
      \centering
        {(g) \\DSTN \cite{DSTN}}
    \end{minipage} 
\caption{\textbf{More qualitative comparison with state-of-the-art methods.}}
\label{fig:comparison2}
\end{figure}
\begin{figure}[t]
\centering
\includegraphics[width=1\linewidth]{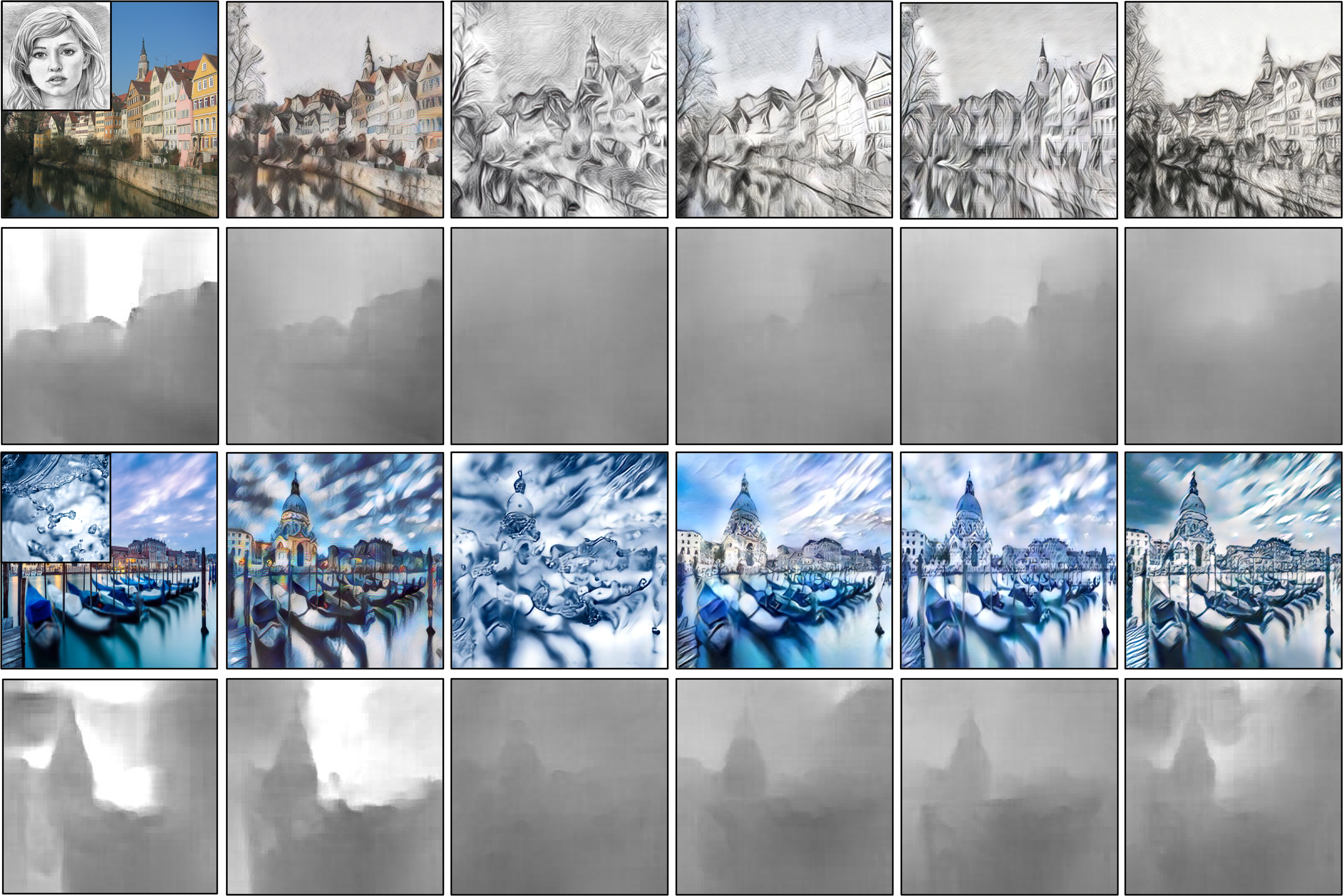}
    \begin{minipage}{0.2\linewidth}
      \centering
        {\footnotesize{(a) \\Style\\Content}}
    \end{minipage}  
    \begin{minipage}{0.15\linewidth}
      \centering
        {(b) \\Ours}
    \end{minipage} 
    \begin{minipage}{0.15\linewidth}
      \centering
        {(c) \\WCT \cite{li2017universal}}
    \end{minipage} 
    \begin{minipage}{0.15\linewidth}
      \centering
        {(d) \\AdaIN \cite{huang2017arbitrary}}
    \end{minipage} 
    \begin{minipage}{0.15\linewidth}
      \centering
        {(e) \\MST \cite{zhang2019multimodal}}
    \end{minipage} 
    \begin{minipage}{0.15\linewidth}
      \centering
        {(f) \\AdaAttN \\\cite{adaattn}}
    \end{minipage} 
\caption{\textbf{Comparison of depth maps.}}
\label{fig:depth}
\end{figure}
\subsection{Qualitative Results}

Figure \ref{fig:arbitrary} shows single style transfer comparison results.
The first image of each set represents a content image. Other images are resulted from seven different styles respectively, which are shown at right-bottom boxes.
Our proposed method maintained the structure of given content images, while transferring the patterns of style images accurately.
Different compositions of target palettes could diversify style transfer results. 

Figure \ref{fig:comparison1} shows single style transfer compared to previous prominent works \cite{huang2017arbitrary,gatys2016image,li2017universal,zhang2019multimodal,adaattn}. 
Our method preserved the detailed structure of content images, while other methods failed (\eg the first row of AdaIN(d) and WCT(f)). 
The performance gap between the proposed method and others was significantly large especially when we used complex-structured content images (\eg a large number of houses in the fourth row) or relatively simple style images (\eg uniform color and texture in overall of the fifth row). 
While other methods were hard to recognize the image contents, our method retained a single object given as a content image.
Moreover, our method fully reflected rich styles with few artifacts, whereas Gatys \etal(e) produced insufficient stylized images on the third row, and AdaIN(d) included halo artifacts near the bridge on the third row. AdaAttN(h) \cite{adaattn} shows fairly good results except failures in color preservation in the third row and content preservation in the firth row.

Figure \ref{fig:comparison2} represents style transfer results compared with the latest works \cite{sheng2018avatar,aams,park2019arbitrary,efanet,DSTN}.
Similar to Figure \ref{fig:comparison1}, our network produces plausible results maintaining the structure of the content.
The proposed method shows fewer repetitions of undesirable patterns (\eg the first row of Avatar-Net(d) and SANet(f)) and enables recognition of the content even after style transfer. Most methods tend to stand on one side in terms of the trade-off between content preservation and pattern reproduction. Because both qualities are needed for successful style transfer, quantifying each side and balancing has to be considered in future works.

We tested five arbitrary end-to-end models and estimated depth map of $1,652$ images, a combination of $28$ content images and $59$ style images, to measure the remaining degree of content's structure. As Liu \etal mentioned in \cite{liu2017depth}, depth map preservation helps satisfactory stylization by retaining the layout of images. We used the WaveletMonoDepth \cite{wavelet-monodepth} to estimate the depth map from a single image. Figure \ref{fig:depth} represents qualitative comparison of depth maps. Each depth map in the second-fourth row of (a) is obtained from a given content image. Higher values(bright regions) mean a farther area from the camera. Maintaining depth values means that the distance between objects and the camera remained after the stylization. It allows to successfully recognize the original objects and their relationships as before. Our method preserves key structure after style transfer without any guidance of information, while other methods fail. These results show that the proposed approach based on a content-focused mechanism has robustness in retaining the layout.

 \begin{table*}[t]
 	\centering
 	\setlength{\tabcolsep}{10pt}
 	\caption{\textbf{Qualitative comparison of depth map error.} Lower numbers indicate better results. The best results are written in boldface.}
 	\label{table:deptherror}
 	\vspace{-3mm}
 	\begin{tabular}{|l|c|c|c|c|c|}
 		\hline
 		 &  Ours & WCT \cite{li2017universal} & AdaIN \cite{huang2017arbitrary} & MST \cite{zhang2019multimodal} & AdaAttN \cite{adaattn}\\ 
 		\hline
 		MAE & \textbf{0.463} & 0.572 & 0.547 & 0.492 & 0.498\\ 
 		\hline
 		RMSE & \textbf{0.571} & 0.688 & 0.665 & 0.601 & 0.622\\ 
 		\hline
 	\end{tabular}
 \end{table*}

\begin{figure}[t]
    \begin{minipage}[b]{1.0\linewidth}
        \centering
        \includegraphics[width=1.0\linewidth]{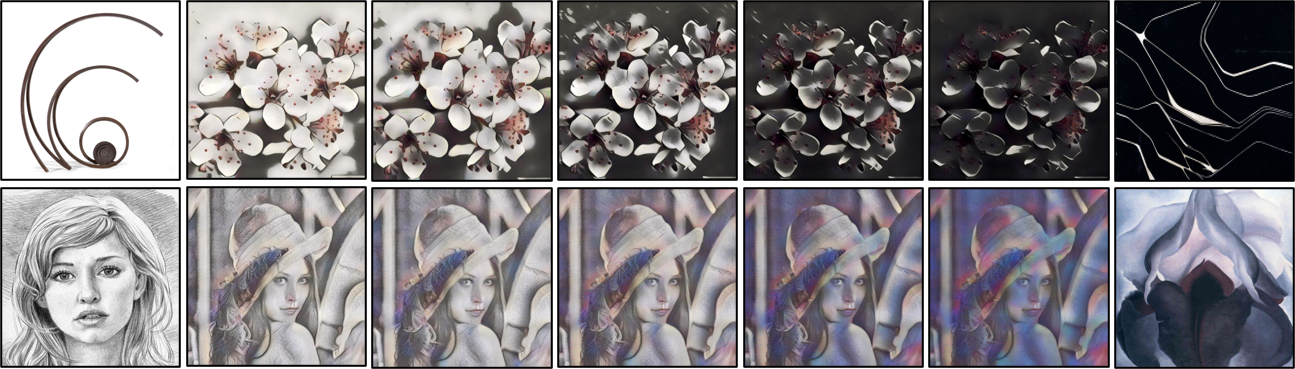}
    \end{minipage}
  \caption{\textbf{Style interpolation with two different pairs.} Linear blending between two style images in the first-last columns. }
 \label{fig:interpolation}
\end{figure}
\subsection{Quantitative Results}

For style transfer problems, quantitative evaluation on content preservation has not been widely progressed. Although Cheng \etal{} \cite{cheng2019structure} attempted to compare the structure consistency via edge detection and depth structure, a single pair of images was used for the evaluation. In this paper, we measured errors between depth maps of content and stylized images using MAE (Mean Absolute Error) and RMSE (Root Mean Square Error) based on large test data. For $1,652$ pairs of content and style image, denoted as $n$, MAE and RMSE can be calculated as follows:
\begin{equation}\label{eq:MAE}
\begin{split}
        MAE = \frac{\sum_{i=1}^{n} |D_{cs} - D_{c}|}{n}, 
        RMSE = \sqrt{\frac{\sum_{i=1}^{n} {(D_{cs} - D_{c})}^2}{n}},
\end{split}
\end{equation}
where $D_{cs}$ and $D_{c}$ denote the depth map of the final stylized image and a given content image, respectively. As shown in Table \ref{table:deptherror}, we obtained the lowest values meaning that layouts of the content image remained than other methods including state-of-the-art methods. This result stems from the content-focusing mechanism of the proposed AC module. The inner procedure of the module relates the content to the first-stylized image and the outer procedure stores up each transferred image. These architectural traits allow the stylization of preserving the overall layout without any guidance of depth or edge information, which is used in \cite{cheng2019structure}.

\subsection{Applications}
Figure \ref{fig:interpolation} represents style interpolation results between the first and last column as given style images. We apply style transfer with given style images and interpolate feature maps of stylized images by controlling their weights.

\begin{figure}[t]
    \begin{minipage}[b]{1.0\linewidth}
		\centering
		\includegraphics[width=1.0\linewidth]{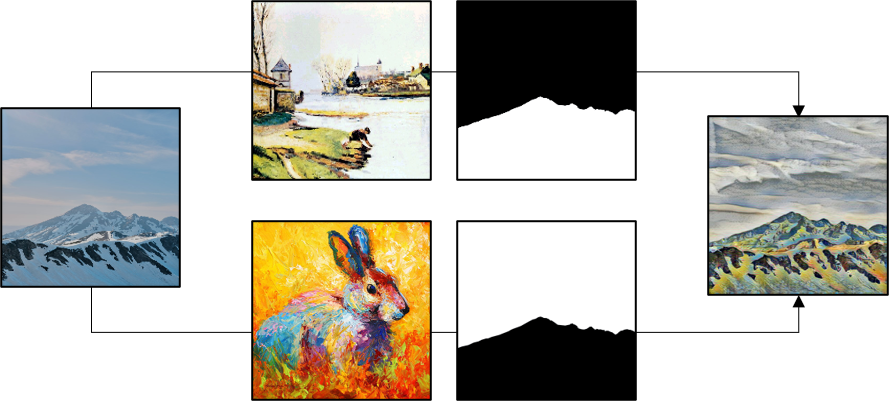}
	\end{minipage}
	\begin{minipage}[b]{0.24\linewidth}
		\centering
		(a) Content image
	\end{minipage}
	\begin{minipage}[b]{0.24\linewidth}
		\centering
		(b) Style images
	\end{minipage}
	\begin{minipage}[b]{0.24\linewidth}
		\centering
		(c) Masks
	\end{minipage}
	\begin{minipage}[b]{0.24\linewidth}
		\centering
		(d) Stylized image
	\end{minipage}
\caption{\textbf{Spatial control}.}
\label{fig:spatial_control}
\end{figure}

Figure \ref{fig:spatial_control} shows a spatial control result using two different style images. Without additional training, the network can produce a stylized image according to desired regions.

\section{Conclusion} \label{sec:con}
Conventional style transfer methods easily lose details of content images and make artifacts in the course of style transfer.
To solve these problems, we propose a novel style transfer method based on a target palette using feature clustering.
We present two main component, namely target palette and attention coloring that plays an important role for style transfer with few artifacts while maintaining the detail of the content.
Our method can produce diverse results using different numbers of clusters. 
Moreover, multiple style input images can be used to generate style-mixed results.
Experimental results demonstrate that our method qualitatively and quantitatively outperforms state-of-the-art arbitrary style transfer methods.
Insufficient transition in complex style patterns should be improved in future work.


\bibliographystyle{spmpsci}      
\bibliography{egbib}   


\end{document}